# Toward Controllable Catalyst Inverse Design via Large-Scale Autoregressive Pretraining

*Dong Hyeon Mok*[1], *Jonggeol Na*[2,3,4*] *and Seoin Back*[4,5,6,7*]

[1]Department of Chemical and Biomolecular Engineering, Institute of Emergent Materials, Sogang University, Seoul 04107, Republic of Korea

[2]Department of Chemical Engineering and Materials Science, Ewha Womans University, Seoul 03760, Republic of Korea

[3]Department of Chemical Engineering, Graduate Program in System Health Science and Engineering, Ewha Womans University, Seoul, 03760, Republic of Korea

[4]Institute for Multiscale Matter and Systems (IMMS), Ewha Womans University, Seoul 03760, Republic of Korea

[5]KU-KIST Graduate School of Converging Science and Technology, Korea University, Seoul 02841, Republic of Korea.

[6]Department of Integrated Energy Engineering, Korea University, Seoul 02841, Republic of Korea

[7]Center for Hydrogen and Fuel Cells, Korea Institute of Science and Technology(KIST), Seoul 02792, Republic of Korea

AUTHOR INFORMATION

**Corresponding Author**

E-mail: sback@korea.ac.kr (SB), jgna@ewha.ac.kr (JN)

**Abstract**

Inverse design of heterogeneous catalysts remains challenging because catalyst surfaces exhibit substantial structural complexity with coupled surface-adsorbate interactions across a vast chemical space that is difficult to explore efficiently through conventional screening alone. Although machine learning-based high-throughput screening has accelerated catalyst discovery, its efficiency inevitably declines as the search space grows, motivating the development of generative models that can directly construct catalysts with target properties. Here, we present a conditional catalyst generative model based on the Generative Pretrained Transformer architecture with a numerical embedding layer that enables the generation of catalyst structures conditioned on both categorical and continuous properties within a single autoregressive framework. The model was pretrained on 133 million catalyst structures and subsequently fine-tuned on approximately 460,000 optimized structures with associated categorical properties and binding energies for conditional generation. The resulting model achieved 98% structural validity, 95% optimization validity, and high categorical condition fidelity, with a 93 % joint match rate for adsorbate type and composition. For binding energy conditioning, the match rate of approximately 20% represents a four-fold improvement over the baseline training distribution, and the generated distributions shift systematically toward the target values, enabling a 1.5 to 4-fold improvement in screening efficiency for reaction-targeted catalyst discovery without additional fine-tuning. Furthermore, targeted fine-tuning on out-of-distribution catalyst domains, including oxide surfaces and single atom catalysts, demonstrated that the model can serve as a practical pretrained backbone that adapts effectively with limited data, consistently outperforming models trained from scratch. These results show that large-scale autoregressive pre-training, combined with explicit property conditioning, provides a practical route toward controllable catalyst generation and accelerated catalysts discovery.

## 1. Introduction

The search for high-performing materials has long been a central goal in chemistry and materials science. Early materials discovery largely relied on trial-and-error approaches, in which human intuition guided repeated cycles of synthesis and evaluation[1]. As experimental and computational techniques advanced, increasingly efficient strategies were developed to explore broader chemical space. Among these, machine learning (ML) has emerged as a major methodology in the field by enabling the prediction of material properties from trained relationships between chemical and geometric information, often without the need for explicit experiments or high-cost simulations[2-4].

Although ML-based approaches are not yet sufficiently accurate to fully replace conventional materials design strategies, they have proven highly effective as screening tools. In particular, their ability to rapidly explore large chemical space allows them to filter out unpromising candidates and prioritize materials that are more likely to satisfy a target property range[5-7]. This paradigm, referred to as high-throughput screening, has been widely adopted across materials discovery and has, in many cases, contributed directly to the identification of experimentally relevant compounds. However, high-throughput screening also has an inherent limitation. As the target search space expands, the number of candidates that must be evaluated grows, eventually increasing the computational cost to the point that the inefficiency of exploration becomes unavoidable[8-10].

To address this limitation, inverse design strategies have been proposed. Rather than exhaustively evaluating all candidates within a predefined chemical space, inverse design aims to directly generate or iteratively optimize materials toward desired properties[11, 12]. Generative models have played an important role in this context from an early stage, and a variety of architectures including variational autoencoders (VAEs), generative adversarial networks (GANs), diffusion models and flow-matching models have been adapted for materials generation[13-16]. In inorganic materials science, these methods have shown promising performance, particularly for relatively simple crystal systems. For example, MatterGen developed by Zeni et al.[17] demonstrated that a diffusion model trained on inorganic crystals containing fewer than 20 atoms could generate physically valid, stable, unique and novel crystal structures with a 40% rate. More recently, a growing body of work has begun to extend generative modeling to inorganic heterogeneous catalysts, which present a considerably greater challenge due to their large structural complexity and the presence of coupled surface-adsorbate configurations[18].

This challenge was partially addressed by the Catalyst Generative Pretrained Transformer (CatGPT), an autoregressive catalyst generative model based on the GPT-2 architecture, in which catalyst structures were represented as token sequences and generated in an autoregressive manner[19, 20]. CatGPT produced valid catalyst structures with a structural validity of 92% and further demonstrated that fine-tuning could steer the generated structures toward the binding energy distributions of the fine-tuning dataset. Building on these capabilities, we proposed an active learning framework in which iterative fine-tuning of the model enabled the selective generation of catalysts with target properties at a rate of 50%, and the resulting catalyst candidates were subsequently validated experimentally[21]. However, CatGPT operates as an unconditional generative model that cannot directly generate catalyst structures conditioned on target properties. Achieving property-targeted generation therefore required additional task-specific fine-tuning for each new target, limiting its practicality as a true inverse design tool. Moreover, extending autoregressive generation to continuous properties such as binding energy posed a fundamental architectural challenge, as the standard transformer embedding layer is designed to process discrete tokens and provides no mechanism for injecting scalar numerical values into the generation process.

In this work, we resolve this challenge by introducing a numerical embedding layer directly integrated into the self-attention mechanism of the GPT-2 architecture, enabling simultaneous conditioning on both categorical properties (adsorbate type, composition) and continuous properties (binding energy) within a single autoregressive framework. Categorical condition tokens were placed prior to the structure tokens during training, and this architectural extension allows the model to jointly process tokenized structural information and continuous numerical features without any task-specific fine-tuning. The model was first pretrained on 133 million catalyst structures to establish a broadly transferable backbone that captures diverse catalyst chemistry across a wide chemical space, providing the scale necessary to achieve high categorical condition fidelity and enable adaptation to new catalyst domains. The pretrained model was then fine-tuned on approximately 460,000 optimized structures to guide generation toward configurations closer to the ground state[22]. With this approach, the model achieved approximately 98% structural validity, 95% optimization validity and 93% joint match rate for categorical conditional generation. For binding energy conditional generation, the match rate was approximately 20%. The binding energy match rate of approximately 20% represents a four-fold improvement over the baseline OC20 training distribution, which showed an average match rate of approximately 5%. Moreover, the binding energy distributions of the generated catalysts shifted clearly toward the target conditions, enabling a 1.5- to 4-fold improvement in screening efficiency for reaction-targeted catalyst discovery. Furthermore, fine-tuning on out-of-distribution domains, including oxide surfaces and single-atom catalysts, demonstrated that the pretrained model serves as a practical backbone that adapts effectively with limited data, consistently outperforming models trained from scratch. These results demonstrate that our conditional generative model provides a practical route toward controllable catalyst inverse design and accelerated catalyst discovery.

## 2. Results

### 2.1. Generative Model for Heterogeneous Catalysts

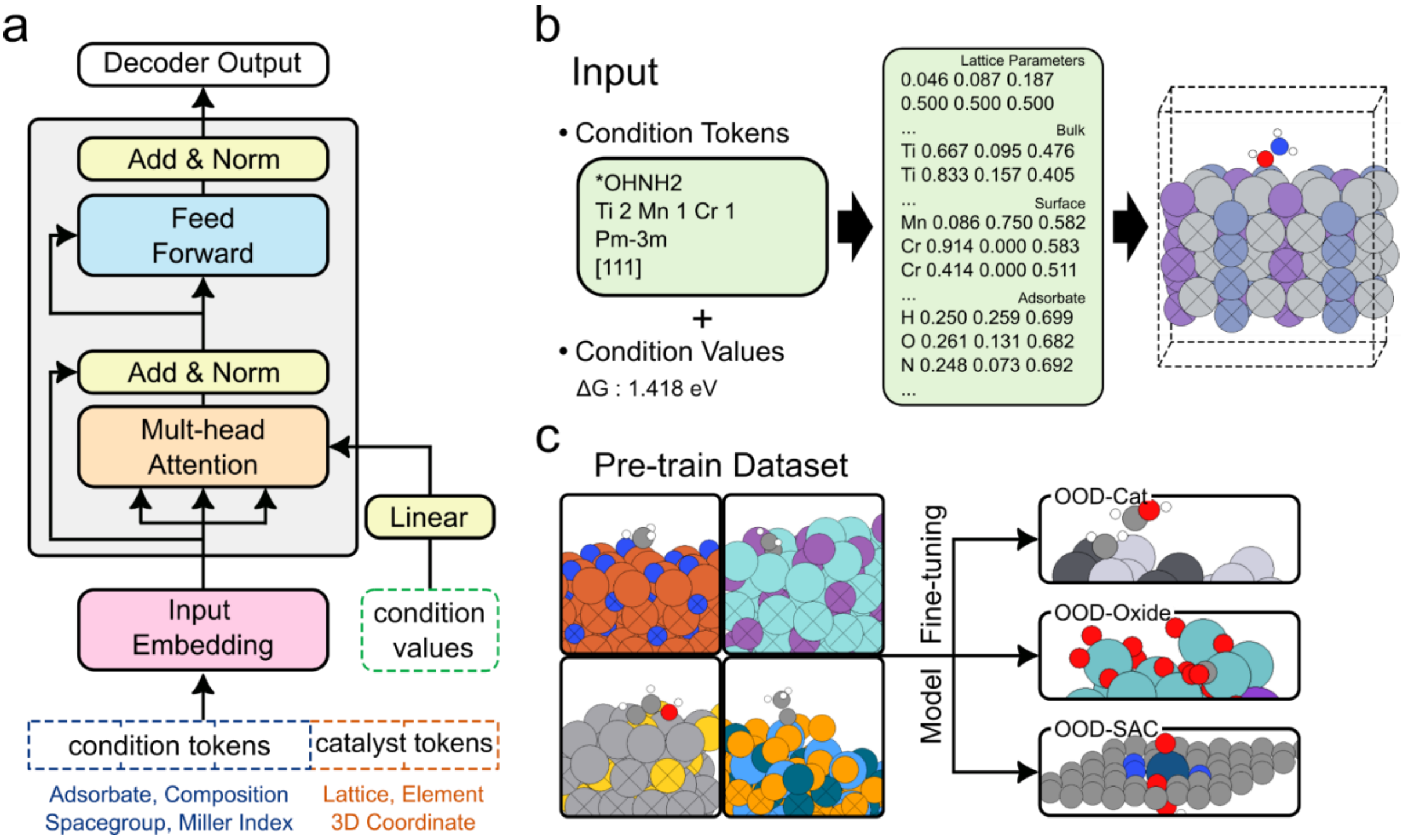

**Figure 1.** (a) Schematic of training the CatGPT model with categorical condition tokens and continuous condition values via an additional embedding layer. (b) Representative example of conditional catalyst generation through a string representation. (c) Fine-tuning the pretrained model for out-of-distribution catalyst generation.

We employed a two-step training process for the catalyst generative model. In the first stage, the model was pretrained on approximately 133 million catalyst structures (OC20-S2EF dataset) represented as strings to learn the syntax of catalyst string representations, token-level relationships, and global geometric patterns[19, 22]. Although this dataset includes structures spanning pre-, intermediate-, and post-optimization states, and is therefore not ideally suited for directly generating relaxed configurations, its large-scale enables the model to robustly capture the semantic meaning of individual tokens and their interdependencies. This large-scale pretraining establishes a strong foundation for generalizable and stable generative behavior. In the second stage, the pretrained model was fine-tuned using approximately 460,328 optimized catalyst structures (OC20-IS2RE dataset). This fine-tuning step biases the generative distribution toward energetically relaxed and physically stable configurations while preserving the diversity and robustness acquired during pretraining.

To evaluate generative performance, we adopted three established metrics: structural validity, uniqueness and novelty. Structural validity assesses physical plausibility through checks on interatomic distances and cell volume. A structure is considered physically valid if all interatomic distances are 0.5 Å or greater and the cell volume exceeds 1.0 $Å^3$.[13] In addition, we introduce optimization validity as a complementary metric to assess whether generated structures can successfully undergo geometry relaxation. This metric is particularly important for catalytic surface structures, where detecting structural pathologies using conventional validity checks alone is challenging compared to bulk crystal systems. Moreover, optimization validity serves as a practical surrogate for stability assessment, which is otherwise difficult due to the necessity of reconstructing the corresponding bulk structures.

Optimization validity was evaluated using UMA (uma-s-1p1), a machine learning force field (MLFF) for material optimization, including catalyst surfaces[23]. A generated structure was deemed to have failed optimization if the maximum atomic force did not converge below 0.05 eV/ Å within 200 optimization steps. For uniqueness and novelty evaluation, conventional structure matcher approaches implemented in Pymatgen were found to be unsuitable for catalyst surfaces. Instead, similarity was quantified using distances between latent vectors embedded by a crystal graph model in a learned latent space (see Section **3.1.**)[24].

Generative performance was assessed by sampling 10,000 catalysts from each model and comparing three training regimes: 1) CatGPT-2M, a model trained on the 2 million S2EF dataset, 2) CatGPT-2M-FT, a model pretrained on the 2 million S2EF dataset and subsequently fine-tuned on 460,328 optimized structures, and 3) CatGPT-133M-FT, a model pretrained on 133 million catalysts followed by fine-tuning on 460,328 optimized structures. The generation took about 5.037 seconds on average per catalyst, and the results show that CatGPT-133M-FT consistently outperformed the baseline models in structural validity, highlighting the benefits of large-scale pretraining. Notably, fine-tuning with optimized structures increased the likelihood of generating successfully optimized catalysts compared to the CatGPT-2M model and CatFlow, a model based on a flow-matching strategy[25], while maintaining comparable structural validity.

**Table 1.** Structural validity (S. Val) and optimization validity (O. Val) for various models. Optimization validity is the proportion of generated catalysts that satisfy structural validity and converge within a force threshold of 0.05 eV/Å within 200 MLFF steps.

| Model | CatGPT-2M | CatGPT-2M-FT | CatGPT-133M-FT | CatFlow |
|---|---|---|---|---|
| S. Val | 0.9123 | 0.9147 | 0.9805 | 0.9733 |
| O. Val | 0.5691 | 0.8313 | 0.9539 | 0.6970 |

## 2.2. Categorical Conditional Generation

We investigate the feasibility of employing a catalyst generative model for inverse design by incorporating conditional control over the generation distribution based on specified catalyst properties. The conditioning variables considered in this work are classified into two types: categorical and numerical. The categorical properties include adsorbates type, composition, space group, and Miller index, while the numerical property corresponds to the binding energy of the adsorbate.

For categorical conditioning, conditional generation is naturally supported by the GPT architecture, which operates as an autoregressive generative model over discrete tokens. Conditioning on categorical properties is achieved by introducing dedicated tokens representing the desired properties at the beginning of the input sequence, preceding the catalyst string representation. In contrast, numerical features cannot be directly processed by the standard transformer embedding layer, which is designed to map discrete tokens into categorical embeddings. To accommodate numerical conditioning, we extended the model architecture by incorporating an additional numerical embedding module that is directly integrated into the self-attention layers (**Figure 1a**). This modification enables the model to jointly process tokenized structural information and continuous numerical features[26]. Through this architectural extension, we developed a conditional catalyst generation framework capable of simultaneously incorporating categorical constraints, namely adsorbate type, composition, space group, and Miller index, and numerical constraints, represented by binding energy, thereby allowing explicit control over the generative distribution.

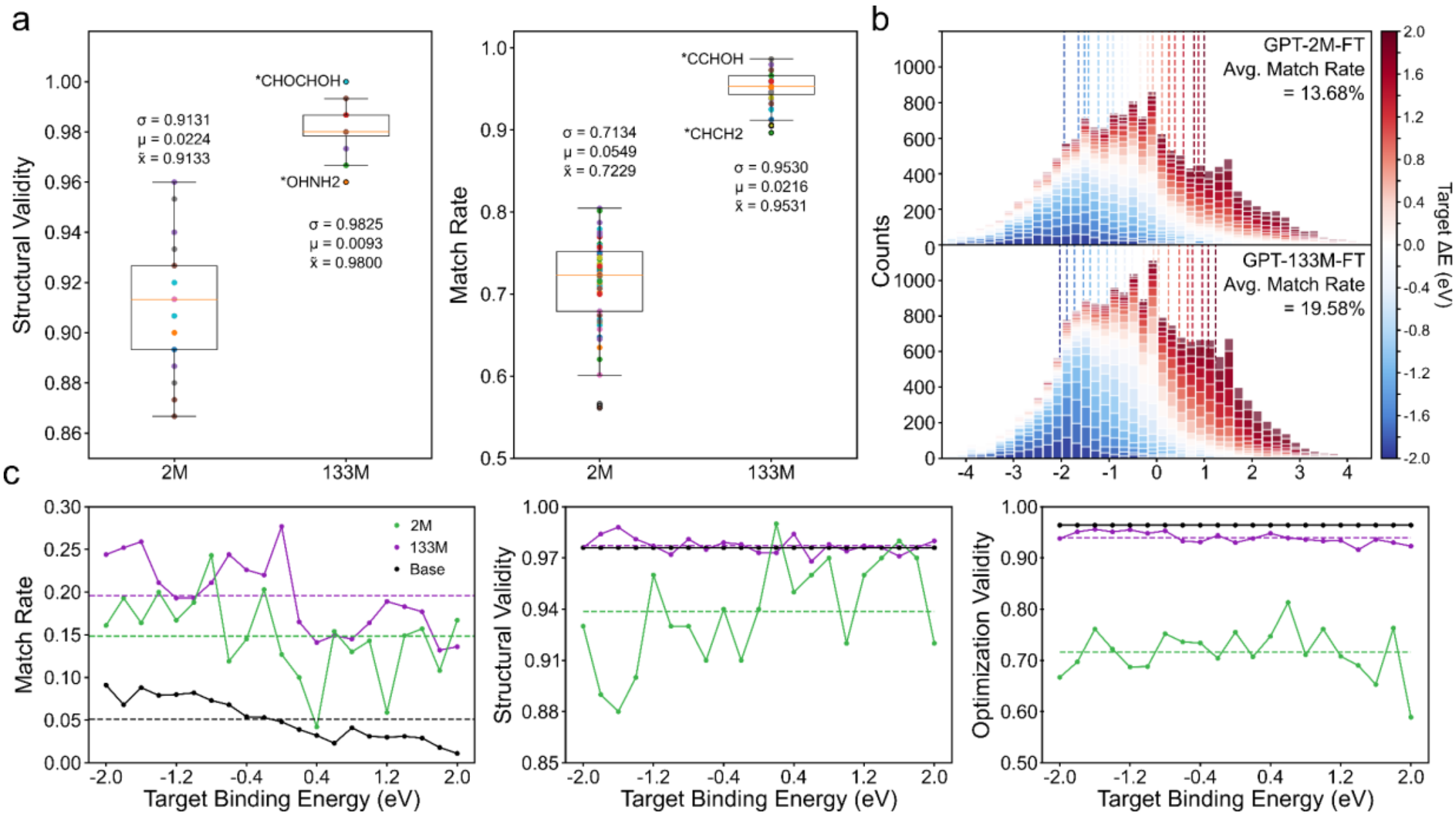

**Figure 2.** (a) Box plots of structural validity and match rate for catalyst structures generated under different adsorbate conditions. (b) Binding energy distributions of structures generated under different binding energy conditions. The upper and lower distributions correspond to CatGPT-2M-FT and CatGPT-133M-FT, respectively. Dashed lines denote the mean of each distribution. (c) Match rate, structural validity, and optimization validity as a function of the binding energy condition. The baseline represents random generation without binding energy conditioning, and the dashed line denotes the mean value.

**Table 2.** Match rates of 10,000 randomly generated catalysts evaluated by whether the adsorbate type and composition of the generated structure match the specified condition tokens. "Both" denotes the joint match rate requiring simultaneous agreement in adsorbate type and composition.

| Match Rate | 2M-FT | 133M-FT |
|---|---|---|
| Adsorbate type | 0.9811 | 0.9833 |
| Composition | 0.2194 | 0.9293 |
| Both | 0.2153 | 0.9293 |

To evaluate the effectiveness of conditional generation limited to categorical constraints, we conducted a text-conditioned sampling in which 150 catalysts were generated per adsorbate using 63 distinct adsorbate types, with random composition tokens specified as additional conditions (**Figure 1b**). The generated structures were then examined to assess compliance with the input adsorbate type and composition tokens. As shown in **Figure 2a**, on average, approximately 95% of the generated catalysts satisfied both categorical constraints while maintaining a high level of structural validity. A more detailed analysis indicates that the probability of generating a catalyst containing the specified adsorbate approached 100% (**Table S1**). This improvement becomes even more evident when the 133M-FT model is compared with the previous 2M-FT model under unconditional sampling. Specifically, after randomly generating 10,000 structures, we evaluated the consistency between the generated structures and their corresponding adsorbate-type and composition tokens. The 2M-FT model achieved a joint match rate of only approximately 22%, whereas the 133M-FT model reached approximately 93% (**Table 2**). This substantial improvement mainly arises from the much higher composition match rate of the 133M-FT model, which is more than four times that of the 2M-FT model. These results likely reflect the benefit of large-scale pretraining, which enables the model to better capture the relationship between composition tokens and physically valid catalyst structures. Considering minor discrepancies between the prescribed bulk composition and the resulting surface composition attributable to variations arising from surface cleavage during catalyst construction, these results demonstrate that the model adheres to the imposed categorical conditions with near-perfect fidelity.

### 2.3. Binding Energy Conditional Generation

We next evaluated the conditional generation performance of the model with respect to numerical binding energy constraints. Binding energy values ranging from –2.0 to 2.0 eV were discretized into intervals of 0.2 eV. For each target binding energy value, 1,000 catalyst structures were generated and subsequently optimized using the UMA framework. The proportion of catalysts whose computed binding energies fell within ±0.2 eV of the target value was then quantified. On average, approximately 20% of the generated catalysts satisfied this criterion. This match rate represents a four-fold improvement over the baseline OC20 training distribution, which showed an average match rate of approximately 5%. Moreover, the binding energy distributions clearly shift toward the specified target values, indicating that numerical conditioning effectively guides the generative distribution toward the desired property regime (**Figure 2b** and **c**).

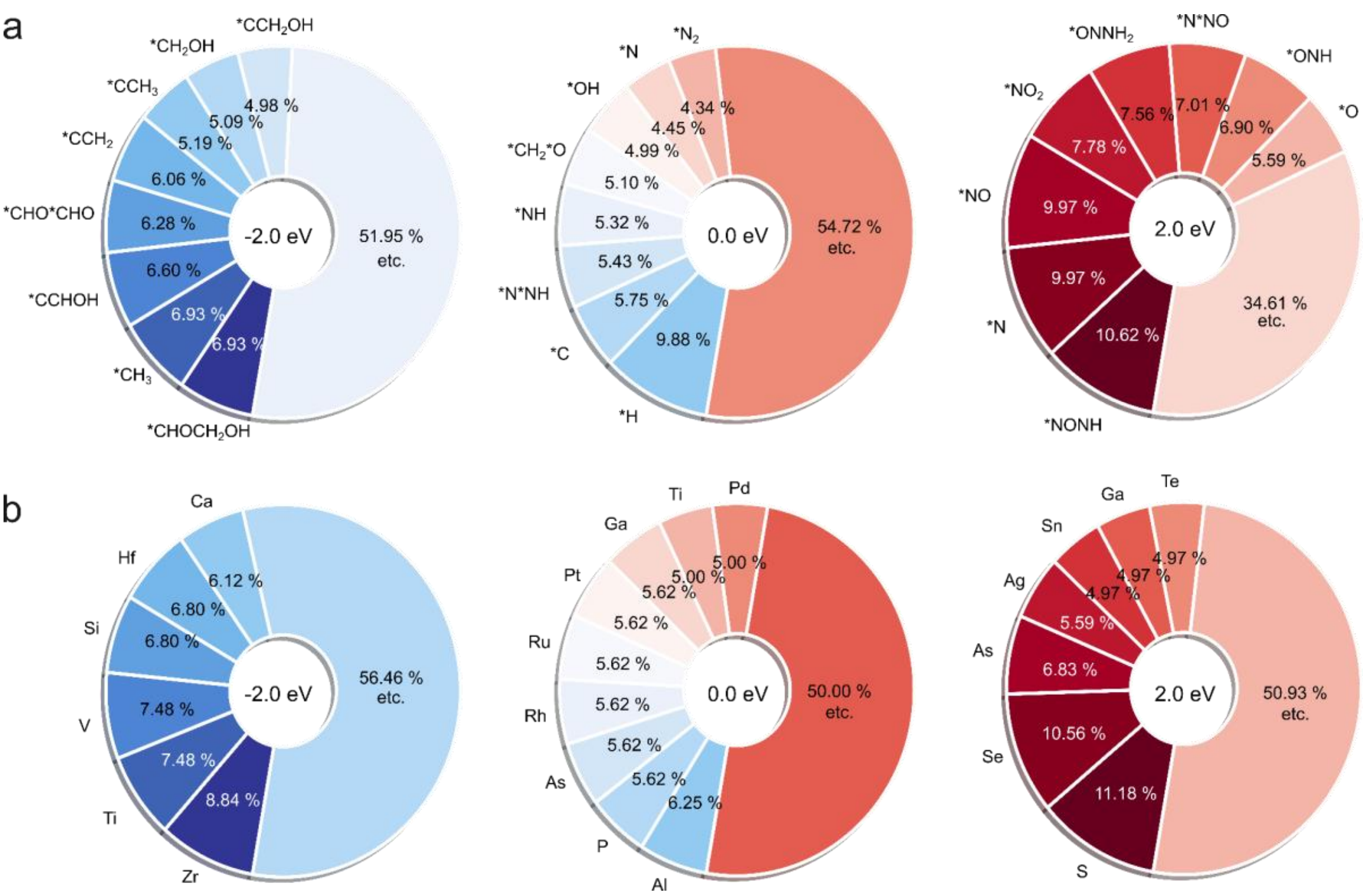

**Figure 3.** (a) Donut plots illustrating the distribution of adsorbate types in generated catalyst structures under binding energy conditions of −2.0, 0.0, and 2.0 eV. (b) Donut plots illustrating the distribution of constituent elements in generated catalyst structures under joint conditioning on binding energy (−2.0, 0.0, and 2.0 eV) and *O adsorbate type.

The influence of property conditioning is further elucidated by simultaneously applying categorical and numerical constraints. As shown in **Figure 3a**, varying the binding energy condition systematically shifted the adsorbate identity distribution of the generated catalysts. This trend indicates that the model has successfully learned the correlation between adsorbate identity and adsorption strength across catalyst surfaces. In addition, as shown in **Figure 3b**, when *O is specified as the adsorbate type in conjunction with strong or weak binding energy targets, the resulting catalyst composition distributions are systematically shifted. In particular, strong binding energy conditions favor the generation of catalysts enriched in oxophilic elements, whereas weak binding energy conditions bias the composition toward oxophobic elements. These results confirm that the model captures meaningful correlations between adsorbate identity, adsorption energetics, and catalyst composition, enabling property-driven catalyst discovery.

Furthermore, we assessed the electrochemical reaction-targeted generation capability of CatGPT-133M-FT model by comparing the property distributions of catalysts produced under joint conditioning on adsorbate type and binding energy with those obtained from adsorbate type-only conditioning. This comparison allowed us to examine whether the model can bias generation toward catalysts with desired adsorbate and adsorption energy distributions, and, consequently, whether it can serve as a practical tool for the discovery of highly active catalysts for specific reactions. Since the model cannot impose conditions on multiple intermediates simultaneously, we focused on reactions representable by a single intermediate or an effective descriptor. Hydrogen evolution reaction (HER) was selected as a representative single-intermediate reaction, with optimal activity corresponding to $\Delta G_{*H} \approx 0$ eV[27]. In addition, the 4-electron oxygen reduction reaction (4e-ORR) was considered because, although it involves multiple intermediates, its activity can be approximated through scaling relations using *OH as a descriptor, where the optimal $\Delta G_{*OH}$ was set to 0.86 eV[28].

As a result, when 1,000 *H and *OH adsorbed catalysts were generated without binding energy conditions, the match rate of catalysts whose Gibbs free energies fell within 0.2 eV of the optimal values for $\Delta G_{*H}$ and $\Delta G_{*OH}$ were 19.64% and 3.51%. The relatively high match rate of $\Delta G_{*H}$ is attributable to the less catalyst-sensitive nature of *H binding. When binding energy conditions were applied, these rates increased to 33.44% and 13.39%, respectively (**Figure S2**). These results indicate that the efficiency of screening catalysts with the desired adsorbate and binding energy can be improved by approximately 1.5- to 4-fold, depending on the adsorbate type, without additional fine-tuning. Taken together, these results demonstrate that the proposed conditional generation framework can serve as an effective inverse design tool for promising catalyst screening.

### 2.4. Application as Foundation Model

**Table 2.** OOD datasets and the corresponding structural validity (S. Val), optimization validity (O. Val), and match rates for adsorbate type (Ads.), composition (Comp.) and both obtained from fine-tuned models on each dataset. "Scratch" denotes models trained directly on the target dataset without pretraining.

| Base Model | OOD-Dataset | S. Val | O. Val | Match Rate | | |
|---|---|---|---|---|---|---|
| | | | | Ads. | Comp. | Both |
| 133M-FT | Cat | 0.8915 | 0.8368 | 0.9990 | 0.9378 | 0.7378 |
| | Oxide | 0.6686 | 0.7226 | 0.9726 | 0.2052 | 0.2051 |

| | SAC | 0.8840 | 0.9471 | 0.1118 | 0.0616 | 0.0615 |
|---|---|---|---|---|---|---|
| 2M-FT | Cat | 0.9362 | 0.8995 | 0.9954 | 0.2176 | 0.2163 |
| | Oxide | 0.7717 | 0.7452 | 0.8431 | 0.0479 | 0.0433 |
| | SAC | 0.9961 | 0.9697 | 0.0000 | 0.0000 | 0.0000 |
| Scratch | Cat | 0.0105 | 0.6000 | 0.9474 | 0.2105 | 0.2105 |
| | Oxide | 0.0095 | 0.5429 | 0.0857 | 0.1809 | 0.0288 |
| | SAC | 0.0000 | - | - | - | - |

We have demonstrated that the model can reliably generate valid catalyst structures within the training data distribution and that this distribution can be effectively modulated through conditional inputs. Building on this capability, we further explore whether the model can function as a foundation model that facilitates the generation of catalyst structures beyond the trained distribution through targeted fine-tuning.

We employed three out-of-distribution (OOD) datasets that exhibit progressively larger deviations from the original training distribution (**Figure 1c**). The first category consists of heterogeneous metal alloy surfaces that were not included in the training data but remain within a similar structural and chemical domain. These datasets contain unseen catalyst surfaces, corresponding to OOD-cat splits in the OC20 benchmarks[22]. The second category comprises oxide surfaces that are entirely absent from the training set, specifically the OC22 IS2RE validation dataset. In contrast to OOD-cat, this dataset involves elemental compositions, bonding environments, and surface chemistry rules that are fundamentally different from those encountered during training. This dataset is referred to as OOD-oxide[29]. The third category, denoted as OOD-SAC, consists exclusively of single-atom catalysts, for which the geometric and coordination environments, and constituent elements differ substantially from those of heterogeneous alloy surface catalysts, representing the most significant distribution shift from the training set (**Figure S1a**). Details about the OOD datasets and latent space embeddings can be found in **Supplementary Note 1**.

We separately fine-tuned the 133M and 2M pretrained models on each OOD dataset and evaluated their effectiveness as foundation models under limited data and strongly shifted distributions relative to the pretraining dataset. Their performance was assessed against models trained from scratch on the same OOD datasets, allowing us to examine how robustly the pretrained models adapt to and generalize across distinct catalyst domains. Because the OOD-oxide and OOD-SAC datasets do not include binding energy labels, the evaluation focused on structural validity and categorical conditional generation performance for both composition and adsorbate type, while numerical conditioning for binding energy was excluded. Regarding structural validity, the 2M-FT model achieved the best performance across all datasets, even outperforming the 133M-FT model (**Table 2**). This may indicate that the 133M-FT is more susceptible to overfitting to structures present in the pretraining data, possibly due to the presence of highly similar structures collected during catalyst optimization trajectories. In contrast, the 133M-FT model consistently outperformed the other models in conditional generation across all OOD systems, suggesting a stronger ability to capture the relationship between condition tokens and catalyst structures represented as strings. Notably, the 133M-FT model exhibited strong conditional generation performance across all OOD domains, whereas models trained from scratch showed substantially poorer performance and, in some cases, failed to generate any valid outputs.

We also observed a clear decline in conditional generation performance as the OOD systems became increasingly distant from the pretraining distribution (**Figure S1b**), indicating that the model, by virtue of its autoregressive formulation, has difficulty adapting to structural token contexts that are rare or entirely absent in the pretraining data. Nevertheless, even on the most distant OOD-SAC dataset, its performance remained meaningful and non-trivial. Overall, these results support the view that the 133M-FT model can serve as a catalyst foundation model, providing practical conditional generative capability even in regimes where the available data are insufficient to train an effective generative model from scratch.

## 3. Discussions

### 3.1. Uniqueness, Novelty, and Catalyst Validity

Evaluating the uniqueness and novelty of generated catalyst structures is essential for assessing both the generative capacity of models and their practical efficiency in accelerating materials discovery. While conventional tools like the "StructureMatcher" module in Pymatgen are highly effective at identifying duplicate bulk crystal structures, and recent strategies, such as the ordered-disordered matching proposed by Zeni et al., successfully address compositional disorder, their applicability to heterogeneous catalyst surfaces is fundamentally limited by the unique modeling paradigms of surface science[17]. Unlike bulk crystals, catalyst surfaces are typically modeled using slab geometries that break periodicity along the z-axis. This introduces specific structural artifacts that traditional coordinate-based matchers fail to account for, such as vacuum padding and varying bottom layer thicknesses, which do not influence active surface chemistry. Because traditional matchers evaluate the entire unit cell uniformly, they erroneously classify identical active surface motifs as distinct structures if they differ only in these trivial aspects.

**Table 3.** Uniqueness and novelty of structures generated by CatGPT-2M-FT and CatGPT-133M-FT relative to the OC20-IS2RE dataset. Both metrics are evaluated using the Structure Matcher (SM) and the Ordered-Disordered Structure Matcher (ODSM). Uniqueness and novelty of proto-structures (Proto) are evaluated using SM without element information.

| | Uniqueness | | | Novelty | | |
|---|---|---|---|---|---|---|
| | SM | ODSM | Proto | SM | ODSM | Proto |
| 2M-FT | 0.9991 | 0.9987 | 0.8051 | 0.9833 | 0.9681 | 0.7248 |
| 133M-FT | 0.9993 | 0.9985 | 0.7919 | 0.9785 | 0.9571 | 0.7068 |

To examine this limitation, we evaluated whether conventional geometry-based matching methods could reliably identify duplicate catalyst structures. Specifically, 10,000 randomly generated catalysts from both the CatGPT-133M-FT and CatGPT-2M-FT models were compared against the relaxed OC20-IS2RE dataset used for fine-tuning. Even under highly permissive StructureMatcher settings (fractional length tolerance *ltol* = 0.3, site tolerance *stol* = 0.5, and angle tolerance *angle_tol* = 10), both models yielded unrealistically high uniqueness and novelty values exceeding 95%. To further isolate geometric diversity from compositional effects, we also evaluated “proto-structures”, in which all atoms were replaced by a single element, and still obtained values of approximately 80% (**Table 3**)[30]. As noted in recent studies on inorganic crystal generation, such near-perfect scores are often inflated. These inflated scores reflect false negatives caused by the algorithm’s oversensitivity to structurally inconsequential variations, such as vacuum padding and varying bottom layer thicknesses, which dictate global cell parameters but do not govern actual surface properties.

To address the limitations of rigid geometric matching, we instead employed a more flexible metric based on continuous latent-space distances. High-dimensional graph embeddings for each catalyst were extracted using EquiformerV2[31], and cosine distances to the nearest neighbor in this embedding space were used to quantify structural and chemical similarity. Uniqueness was defined as the fraction of generated catalysts having no neighbor within the generated set below a given distance threshold, whereas novelty was defined as the fraction of generated catalysts having no neighbor within the fine-tuning set below that same threshold. By systematically varying this threshold, we constructed cumulative distribution function (CDF) curves for both uniqueness and novelty (**Figure 4a** and **b**).

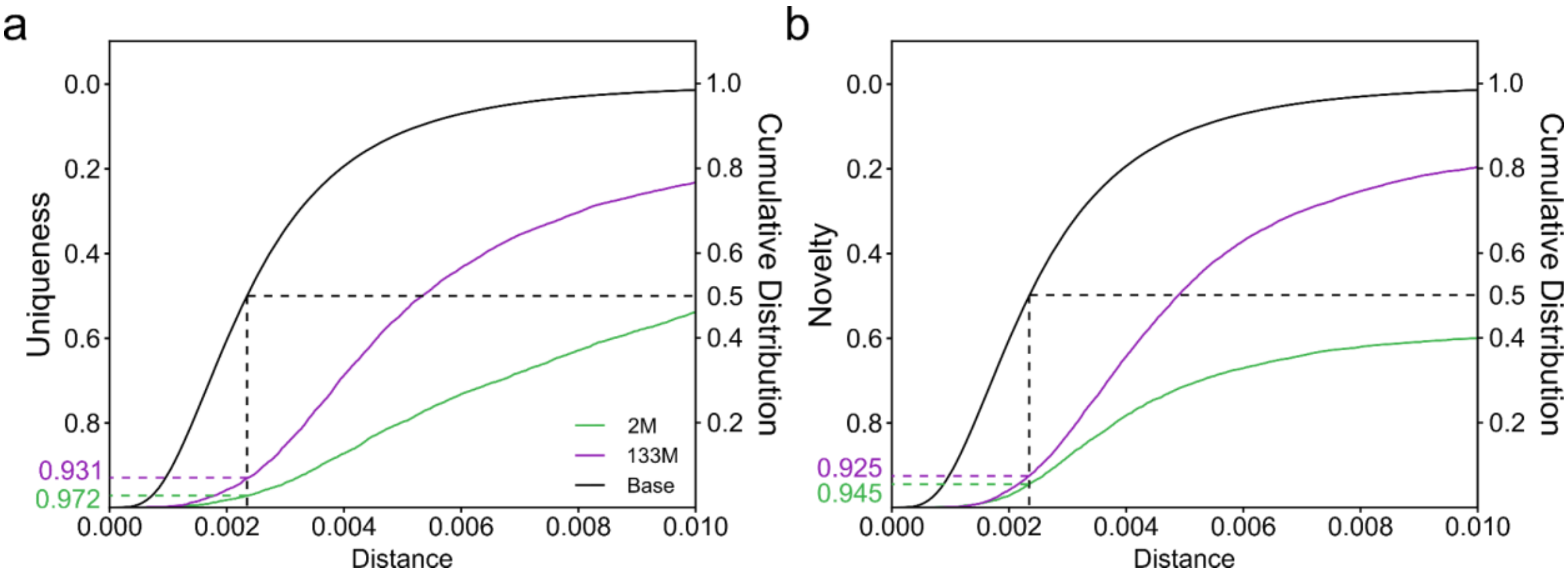

**Figure 4.** Cumulative distribution functions (CDFs) of nearest-neighbor cosine distances in the catalyst latent space. (a) Uniqueness is defined as the fraction of generated catalysts with no other generated catalyst within a given distance threshold, whereas (b) novelty is defined as the fraction with no training set (Base) neighbor within the same threshold. Dashed lines indicate the reference distance at which the baseline training set reaches 50% uniqueness and the corresponding uniqueness and novelty values of generated catalysts at that distance.

Interestingly, both models generated structures that were highly distinct and novel relative to the fine-tuning dataset. Using the median nearest-neighbor distance of the fine-tuning data ($D_{50}$) as a reference threshold, the CatGPT-133M-FT model achieved uniqueness and novelty values of 93.06% and 92.52%, respectively, whereas the CatGPT-2M-FT achieved 97.23% and 94.49%. These results suggest that, although *StructureMatcher*-based evaluations are imperfect, they may still capture some degree of the distinctiveness of generated catalyst structures.

However, in contrast to the similar performance suggested by *StructureMatcher*, the latent space CDFs revealed that the CatGPT-2M-FT model exhibited consistently higher uniqueness and novelty across nearly all distance thresholds. We attribute this, at least in part, to the inclusion of structures that are formally distinct in latent space but cannot be regarded as realistic catalysts due to a lack of catalyst validity. Our previous work reported abnormal structures, such as catalysts with missing atoms or completely broken bonding networks caused by pathological cell sizes. Because such cases are not readily excluded by distance-based structural validity criteria alone, prior studies introduced auxiliary detection models to identify anomalous catalysts. Nevertheless, those approaches depended strongly on the training data distribution and could incorrectly classify genuinely novel catalysts as anomalous. For this reason, we did not adopt that strategy here. As a result, catalysts located far from the ground-truth distribution may have inflated the apparent uniqueness and novelty of the CatGPT-2M-FT. Indeed, we identified 249 catalysts with no close neighbors, either within the generated set or the fine-tuning dataset, even beyond $D_{99.99}$ of the fine-tuning data, and some of these structures were clearly anomalous (**Figure S3** and **S4**).

A remaining challenge is that latent distance-based metrics for uniqueness, novelty and catalyst validity remain inherently ambiguous. Because some catalysts outside $D_{99.99}$ still exhibited chemically reasonable structures, it is not possible to classify all isolated structures as invalid, nor is it straightforward to define a universal distance threshold for novelty or uniqueness. This issue remains an open challenge and highlights the need for improved evaluation metrics capable of more accurately assessing the validity, novelty and uniqueness of catalyst structures produced by generative models.

### 3.2. Challenges and Future Directions

A key limitation of the autoregressive framework and token-based catalyst representation adopted in the CatGPT architecture is that fine-tuning and categorical conditional generation are fundamentally restricted to the predefined vocabulary and token space available during pretraining. As a result, the model cannot be readily fine-tuned on or used to generate truly out-of-domain catalysts containing unseen elements or structural attributes. This limitation arises because Transformer decoder-based models do not learn intrinsic chemical or structural meanings from individual tokens. Instead, they learn statistical relationships among tokens within the training distribution. Consequently, the present model cannot generate catalysts containing new adsorbate types, elements, compositions, space groups or Miller indices that were not encountered during pretraining. Within the scope of Transformer decoder-based architectures, potential strategies to address this limitation include expanding the training dataset to cover a broader range of categorical features and chemical species, or augmenting token embeddings with chemically and structurally informative feature vectors that explicitly encode the underlying properties of each element and structural token.

Another significant challenge is the difficulty of evaluating catalyst stability due to the non-trivial inverse problem of reconstructing the parent bulk structure from a generated surface slab. Because surface structures do not uniquely determine their corresponding bulk phases, reliable stability evaluation remains difficult within a generative framework of this kind. Nevertheless, the current model offers a partial improvement in this regard. Specifically, the string representation of a catalyst includes bulk composition and space group information, which are generated together with the surface structure and can substantially narrow the search space for plausible bulk candidates. In addition, the model outputs Miller indices, which may further facilitate bulk reconstruction through a systematic enumeration strategy. Candidate bulk structures consistent with the generated composition and space group can be enumerated, surfaces can then be cleaved along the predicted Miller indices, and the resulting slabs can be matched to the generated surface structures. Although this approach does not fully resolve the bulk recovery problem, it provides a more practical route toward connecting generated surface catalysts to their underlying bulk structures and, ultimately, toward more reliable stability assessments.

## 4. Conclusions

In summary, we developed a conditional autoregressive catalyst generative model that enables direct generation of catalyst structures under both categorical and continuous property constraints. By combining large-scale pretraining on 133 million catalyst structures with fine-tuning on optimized configurations, the model achieved high structural validity and strong categorical conditional fidelity, while numerical conditioning on binding energy produced clear distribution shifts toward the target values. Furthermore, targeted fine-tuning on out-of-distribution catalyst domains demonstrated that the pretrained model can serve as an effective foundation model for catalyst generation under limited data conditions. Although challenges remain in evaluating the novelty and structural stability of generated catalyst surface structures, these results establish conditional autoregressive generation as a practical and scalable framework for catalyst inverse design and accelerated catalysts discovery.

## 5. Method

### 5.1 Representations of Catalyst Structures

Catalyst structures can be fully represented by the lattice lengths and angles of the unit cell ($l_1$, $l_2$, $l_3$, $\theta_1$, $\theta_2$, $\theta_3$), atomic symbols ($e_i$) and their respective fractional (or absolute) coordinates ($x_i$, $y_i$, $z_i$) [32]. Prior to this geometric representation, property descriptors are included to provide global context to the LLM: the type of adsorbate ($A$), the chemical compositions ($M$), the space group ($G$) and the Miller indices ($H$).

During the tokenization process, all components are mapped to discrete string tokens within a unified vocabulary space. The adsorbate type ($A$), space group ($G$) and Miller indices ($H$) are each embedded as a single discrete token. The composition (M) is represented as a variable-length sequence consisting of alternating element tokens ($m_j$) and their corresponding stoichiometric number tokens ($v_j$), such that M = ($m_1$, $v_1$ …, $m_k$, $v_k$).

To process continuous spatial data, we employed a coordinate-level tokenization strategy that directly represents each fractional coordinate as a distinct, fixed precision string token (e.g., '0.000', '0.001', …, '1.000'). To ensure that absolute lattice lengths and angles share this identical numerical token space, their original values are uniformly divided by 180. Furthermore, because the sequence of elements inherently embeds structural constraints, the atomic components are not placed arbitrarily but are systematically ordered as bulk atoms, followed by surface atoms and finally adsorbate atoms. Following the methodology of Ref., atoms located in the first layer of the slab, excluding the adsorbates, are classified as surface atoms, while the remainder are treated as bulk atoms, these are separated by the <sep> tokens.

Thus, the complete tokenized sequence (tuple $C$) for representing a catalyst surface consisting of $N$ atoms is configured as follows:

$$C = (A, m_1, v_1, ..., m_k, v_k, G, H, l_1, l_2, l_3, \theta_1, \theta_2, \theta_3, e_1, x_1, y_1, z_1, ..., e_N, x_N, y_N, z_N),$$

### Model Architecture and Training

To generate the 3D catalyst structure represented by the tokenized sequence (tuple $C$), we employed an autoregressive transformer framework based on the GPT-2 architecture[20]. In this framework, the sequence generation is modeled by predicting each subsequent token $t_i$ via a categorical probability distribution conditioned on all preceding tokens $P(t_i|t_{0:i-1})$.

Consequently, the joint probability of generating a complete tokenized structure sequence $x$ is factorized as follows:

$$P(x) = \prod_{i=1}^{n} P(t_i | t_{0:i-1}) \quad (1)$$

Unlike our previous approach, CatGPT, which strictly utilized the baseline generative architecture without task conditioning, the model in this study incorporates a novel continuous property conditioning mechanism[33]. To enable the generation of structure guided by specific continuous values, especially binding energy, a scalar condition value is first projected through a linear layer to form a condition embedding, denoted as $z_c$. This condition embedding is then explicitly integrated into the multi-head self-attention blocks of the transformer. Specifically, the queries ($q_i$), keys ($k_i$) and values ($v_i$) for the attention mechanism are computed by linearly combining the conventional token hidden states ($z_i$) with the shared condition embedding.

$$q_i = z_i W^Q + z_c W^{Q_c} \quad (2)$$

$$k_i = z_i W^K + z_c W^{K_c} \quad (3)$$

$$v_i = z_i W^V + z_c W^{V_c} \quad (4)$$

By modifying the attention matrices in this manner, the model dynamically shifts its structural generation pathways to satisfy the requested continuous constraints.

The model pre-trained on OC20-S2EF-133M employed an architecture with 12 self-attention layers, 12 attention heads and a hidden embedding size of 768. Pre-training was conducted for 600,000 steps using 8 NVIDIA H200 GPUs with a per-GPU batch size of 96. The pre-trained checkpoint was subsequently fine-tuned on the OC20-IS2RE dataset for 10 epochs. The model pre-trained on OC20-S2EF-2M used 12 self-attention layers, 8 attention heads, and an embedding size of 512, and was trained on a single NVIDIA RTX 3090 GPU with a batch size of 144 for 10 epochs, followed by fine-tuning on OC20-IS2RE for an additional 10 epochs.

**Dataset Details**

To pre-train the models for generating string representations of catalyst structures, we utilized the OC20-S2EF dataset, which comprises single point calculated structures sampled from DFT relaxation trajectories with their associated binding energies[22]. For the CatGPT-2M, we employed the predefined 2M subset provided by the OC20 database. For CatGPT-133M, the training data consisted of 133 million structures, encompassing the entirety of the OC20-S2EF dataset comprising heterogeneous catalyst surfaces with adsorbates. Following pre-training, both models were fine-tuned using 460,328 fully optimized catalyst structures from the OC20-IS2RE dataset. This fine-tuning phase yielded the final models designated as CatGPT-2M-FT and CatGPT-133M-FT.

To evaluate the generation capabilities, we constructed an OOD test suite comprising three distinct structural categories; 1) OOD-cat is sourced from the OC20-IS2RE OOD-Cat split to evaluate extrapolation to unseen catalyst surfaces whose slab compositions and geometries are absent from the training sets[22]. 2) OOD-oxide is sourced from the OC22-IS2RE validation in-domain dataset. Unlike the metal-alloy-focused OC20 data, this dataset comprises oxide catalyst surfaces where oxygen is intrinsically incorporated into the slab[29]. 3) OOD-SAC is sourced from a custom dataset of single-atom catalysts (SACs). These structures consist of a 2D carbon shell support featuring a central metal atom coordinated by four non-metal atoms in the primary coordination shell. To ensure structural diversity, these SACs were generated by randomly substituting the central metal and the first-shell coordinating atoms (N, C, O, H) and introducing various heteroatoms into the second coordination shell.

**Metrics Details**

Structural validity was defined to evaluate the physical plausibility of generated catalyst structures. Specifically, a generated structure was considered structurally valid when the minimum interatomic distance within the unit cell was no less than 0.5 Å and the cell volume exceeded 1 $Å^3$, thereby excluding structures with atomic overlap or unphysical cell geometries. Optimization validity was defined as the ratio of generated catalysts corresponding to energetically reasonable configurations that are not excessively distant from a local minimum on the potential energy surface, which serves to identify structurally anomalous catalysts that may not be filtered out by the structural validity criterion alone. A structure satisfies optimization validity if MLFF-based optimization converges to a maximum force below 0.05 eV/Å within 200 steps.

For categorical conditional generation, performance was evaluated for composition and adsorbate type, while space group and Miller index were excluded because their reliable evaluation was not straightforward. The match rate for each categorical property was defined as the ratio of generated structures whose composition or adsorbate type exactly matched the given condition token. In addition, the 'both' metric was defined as the ratio of generated catalysts that simultaneously satisfied both conditions. For continuous conditional generation with respect to binding energy, the energies of generated structures were evaluated using the MLFF. The conditional generation success rate was then defined as the ratio of generated structures whose binding energies fell within 0.2 eV of the given condition value.

## Computational Details

To optimize SAC structures for OOD fine-tuning, spin-polarized DFT calculations were performed using the Vienna Ab initio Simulation Package (VASP, version 5.4.4)[34, 35] with the projector-augmented wave (PAW) method[36] and the revised Perdew–Burke–Ernzerhof (RPBE) exchange–correlation functional within the generalized gradient approximation (GGA)[37]. Geometry optimizations were performed using convergence criteria of $10^{-4}$ eV for total energy and 0.05 eV/Å for atomic forces, with a plane-wave kinetic energy cutoff of 400 eV. Monkhorst–Pack k-point meshes[38] were constructed as $k_1 \times k_2 \times 1$, such that 25 Å < $a_n \times k_n$ < 30 Å (n= 1, 2), where $a_1$ and $a_2$ denote the lattice vector lengths along the x and y direction respectively.

## Data Availability

OC20 dataset can be found in https://fair-chem.github.io/oc20/. OC22 dataset can be found in https://fair-chem.github.io/oc22/. OOD-SAC dataset can be found in https://github.com/SeoinBack/CatGPT

## Code Availability

The code developed in this work and relevant information can be found in Github (https://github.com/SeoinBack/CatGPT).

## Acknowledgements

S.B. acknowledges the support from the National Research Foundation of Korea (NRF) funded by the Korea government (MSIT and MOE) (RS-2025-00513832, RS-2025-02214715, RS-2025-16063688), and the Nano & Material Technology Development Program through the NRF funded by the Ministry of Science and ICT (RS-2024-00448287). S.B. also acknowledges generous supercomputing time from KISTI. This research was also supported by Korea Basic Science Institute (National Research Facilities and Equipment Center) grant funded by the Ministry of Science and ICT (No. RS-2024-00404602).

**Competing Interest**

The authors declare no competing financial interests.

# Supplementary Information for

# "Toward Controllable Catalyst Inverse Design via Large-Scale Autoregressive Pretraining"

*Dong Hyeon Mok[1], Jonggeol Na[2,3,4*] and Seoin Back[4,5,6,7*]*

*[1]Department of Chemical and Biomolecular Engineering, Institute of Emergent Materials, Sogang University, Seoul 04107, Republic of Korea*

*[2]Department of Chemical Engineering and Materials Science, Ewha Womans University, Seoul 03760, Republic of Korea*

*[3]Department of Chemical Engineering, Graduate Program in System Health Science and Engineering, Ewha Womans University, Seoul, 03760, Republic of Korea*

*[4]Institute for Multiscale Matter and Systems (IMMS), Ewha Womans University, Seoul 03760, Republic of Korea*

*[5]KU-KIST Graduate School of Converging Science and Technology, Korea University, Seoul 02841, Republic of Korea.*

*[6]Department of Integrated Energy Engineering, Korea University, Seoul 02841, Republic of Korea*

*[7]Center for Hydrogen and Fuel Cells, Korea Institute of Science and Technology(KIST), Seoul 02792, Republic of Korea*

AUTHOR INFORMATION

**Corresponding Author**

E-mail: sback@korea.ac.kr (SB), jgna@ewha.ac.kr (JN)

## Supplementary Note 1. Latent-Space Embedding Analysis

To assess the distances between structures or distributions for analyzing the uniqueness, and novelty of generated catalysts and the relationships between the training and out-of-distribution datasets, we constructed continuous latent representations of catalyst structures using EquiformerV2. For each catalyst structure, graph-based features were extracted from the internal scalar embeddings of the model during inference using the pre-trained model. Specifically, the input embedding to the final energy prediction block was used, and the graph latent vector was obtained by averaging the scalar node features over all atoms, yielding a fixed length 128-dimensional representation for each structure. These embeddings were used as a chemically and structurally informed descriptor space for numerically comparing distributions.

### 1.1 Visualization

We visualized the relationships among datasets by reducing their dimensionality using Uniform Manifold Approximation and Projection (UMAP). The UMAP reducer was fitted on the latent vectors of the OC20-IS2RE fine-tuning dataset (denoted as the base dataset), and the generated catalyst and out-of-distribution (OOD) datasets were subsequently projected into the same two-dimensional space using the fitted transformation. This procedure enables direct visual comparison of how closely each dataset overlaps with, deviates from, or extends beyond the base data distribution.

## 1.2 Out of Distribution Dataset

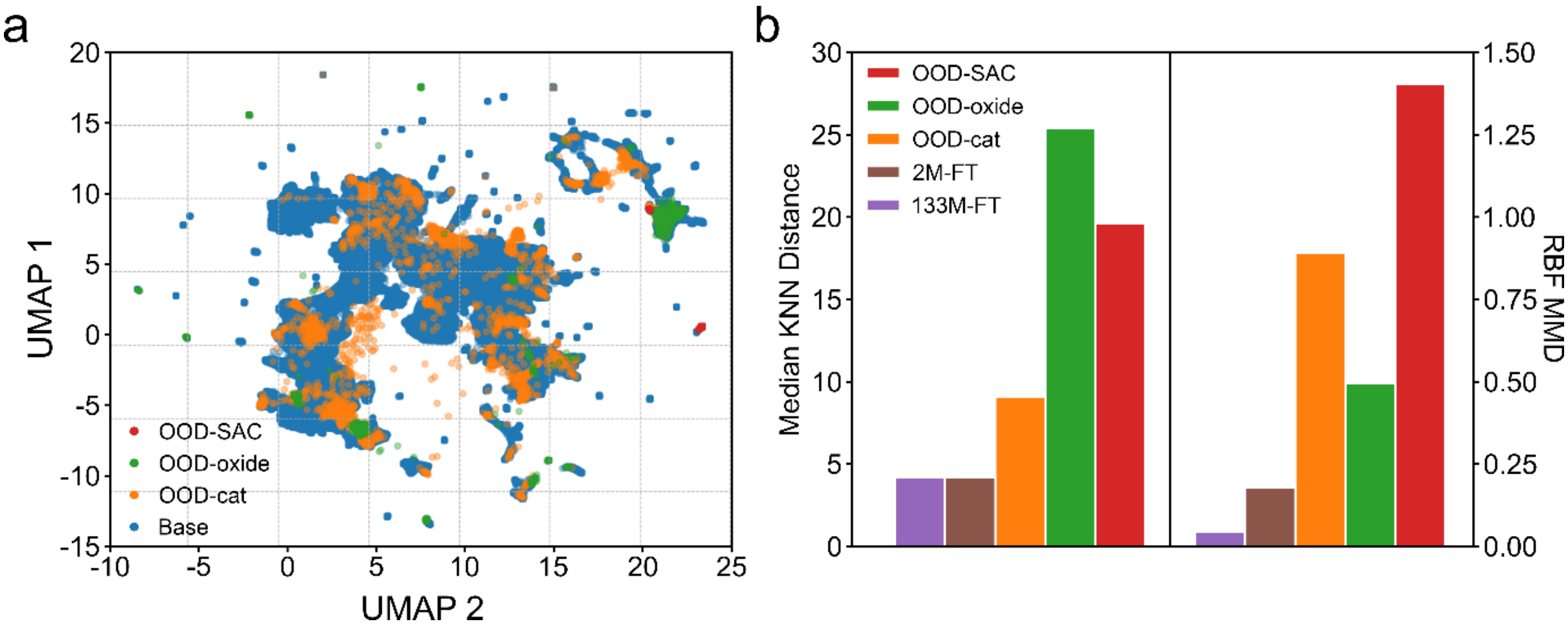

**Figure S1.** (a) UMAP projection of EquiformerV2-derived embedding vectors for the training dataset (Base) and the three out-of-distribution (OOD) datasets. (b) Distributional distances between the training dataset and the five fine-tuning datasets, measured using KNN-based distance and MMD. Greater distances indicate that the fine-tuning dataset is more dissimilar to the training-data distribution.

To quantify the distances between dataset distributions rigorously, we employed two complementary metrics, k-nearest-neighbor (KNN) distance and maximum mean discrepancy (MMD) **(Figure S1)**. Before comparison, all latent vectors were standardized using a Standard Scaler fitted on one reference set and applied consistently to the remaining datasets. For the KNN-based analysis, we computed, for each query structure, the mean Euclidean distance to its five nearest neighbors in the base dataset. The resulting per-structure distances provide a local measure of how far each dataset lies from the reference training distribution. In parallel, we evaluated the global discrepancy between distributions using a radial basis function (RBF) kernel for MMD. Because the base dataset is substantially larger than the comparison groups, repeated random subsampling of the base dataset was performed to match the comparison group size, and MMD values were averaged over multiple repetitions.

### 1.3 Uniqueness and Novelty

We used nearest-neighbor distance distributions trying to analyze uniqueness and novelty. In this framework, uniqueness measures the extent to which generated catalysts are isolated from one another, whereas novelty measures the extent to which they are separated from the base dataset, the OC20-IS2RE fine-tuning dataset. For uniqueness, the nearest-neighbor distance of each structure was computed within the same dataset after excluding self-matches. For novelty, the nearest-neighbor distance from each generated structure to the base dataset was computed. By varying a distance threshold and calculating the fraction of structures with a nearest-neighbor distance below that threshold, we constructed cumulative distribution functions (CDFs) for both uniqueness and novelty related analyses. The median nearest-neighbor distance of the base dataset, denoted as $D_{50}$, was used as a reference point for comparing datasets on a common scale. Lower CDF values at a given threshold indicate a larger fraction of isolated structures, whereas higher values indicate stronger clustering or greater overlap with the reference distribution.

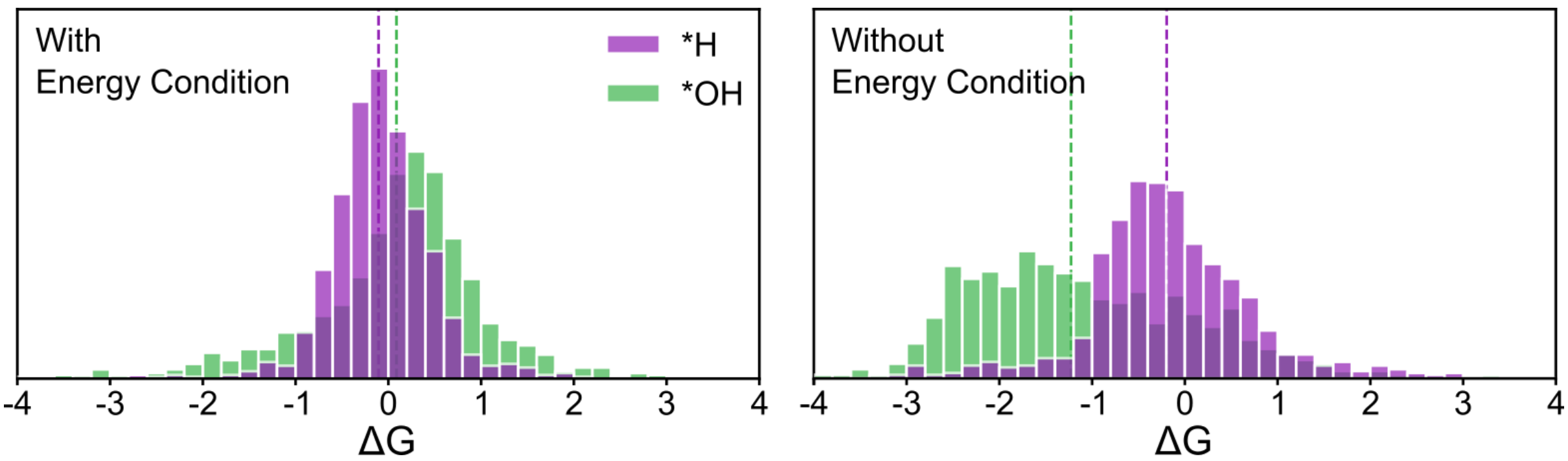

**Figure S2.** *H and *OH adsorbate conditional generation with and without binding energy joint condition. The given binding Gibbs free energy conditions are 0.0 eV and 0.86 eV for *H and *OH, respectively. Dashed lines are average Gibbs free energy of each energy distribution.

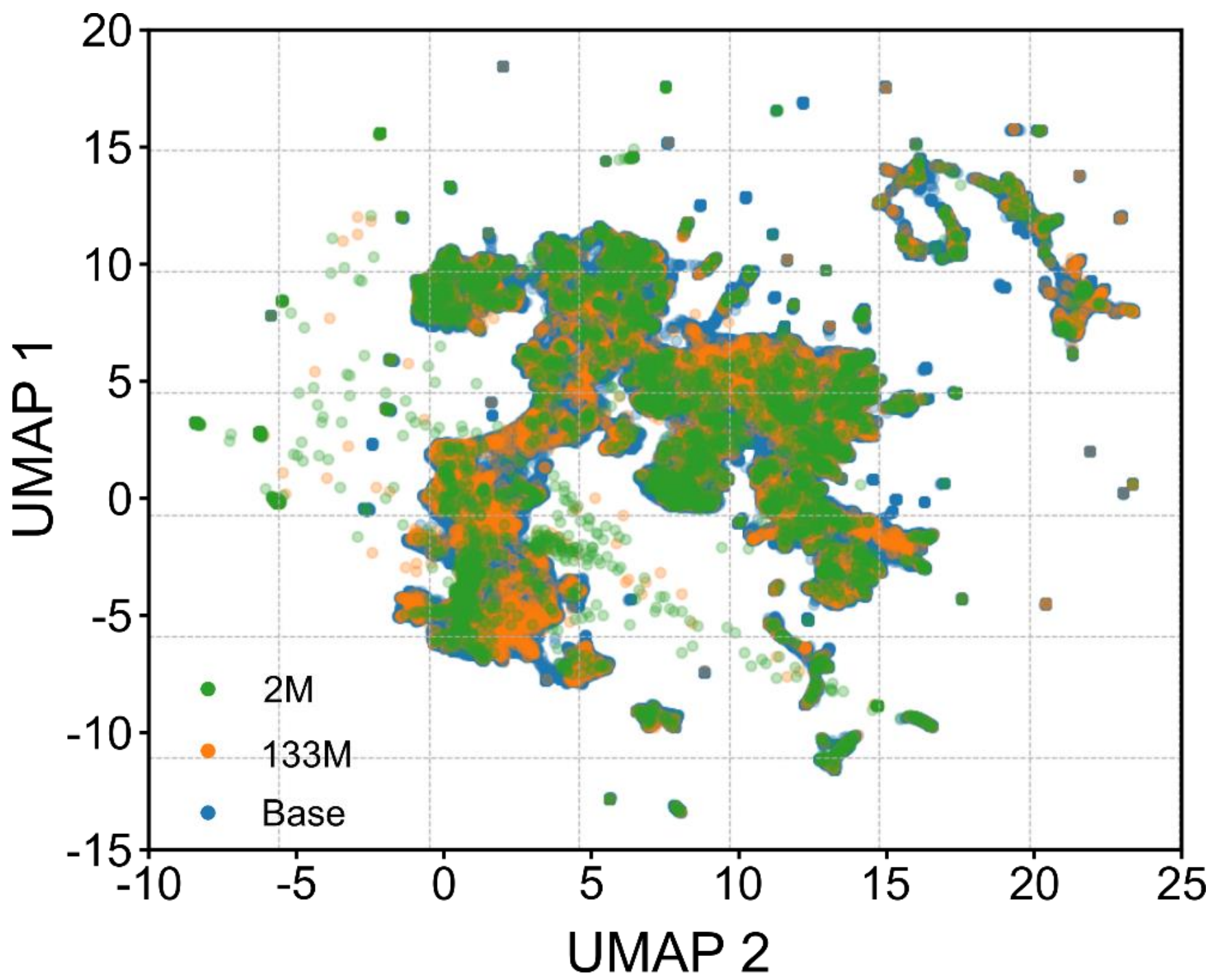

**Figure S3.** Visualization of the embedded vectors of training data (Base) and structures generated from the 133M-FT and 2M-FT models projected onto a 2D space using UMAP.

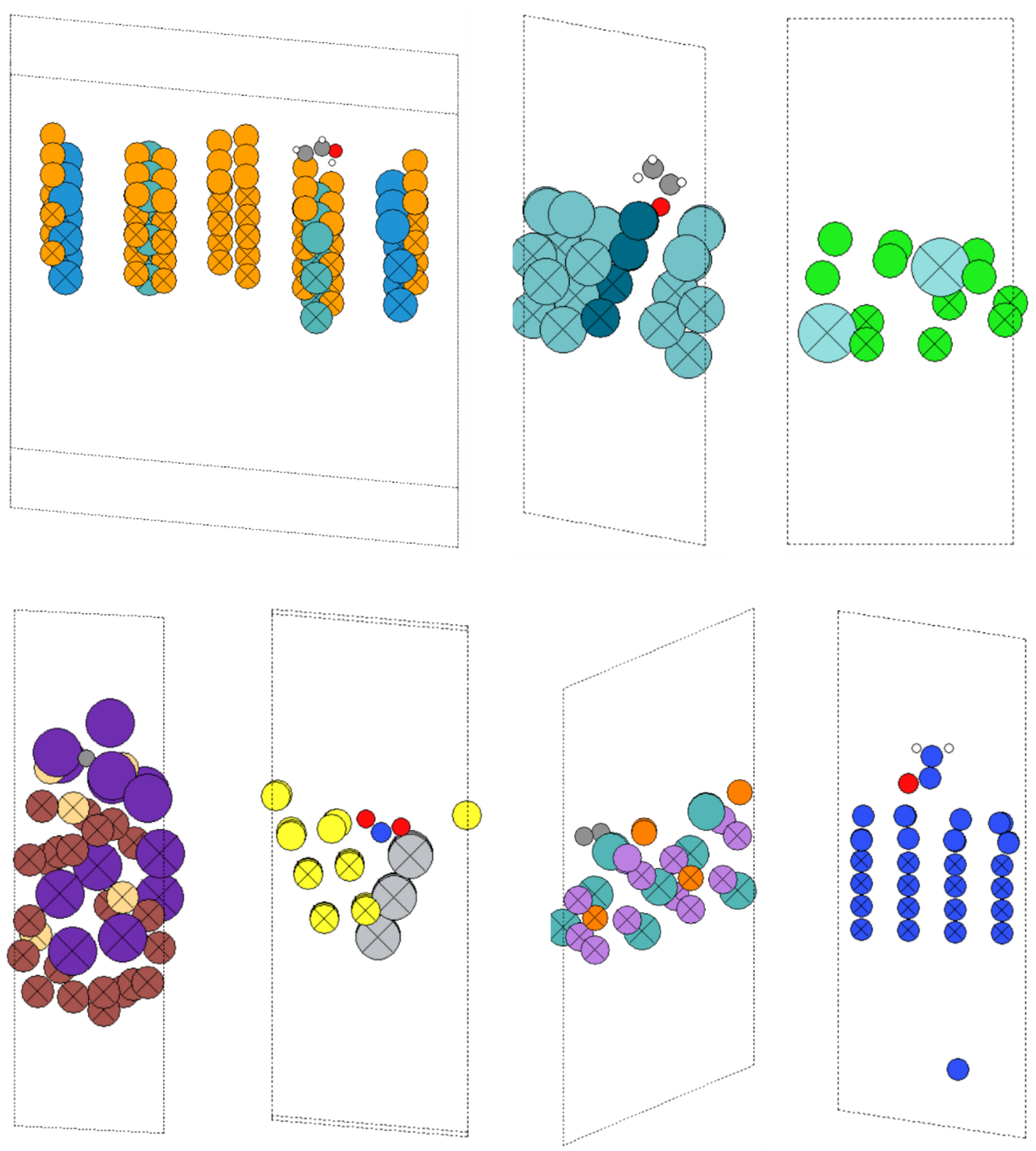

**Figure S4.** Representative latent-space outliers among generated catalyst structures. The structures exhibit anomalous configurations with unrealistic geometries or broken structural motifs.

**Table S1.** Structural validity (S. Val), adsorbate type match rate and composition match rate of structures generated under adsorbate type conditioning. A total of 150 structures were generated per adsorbate type.

| Adsorbate Type | S. Val | Ads. Match | Comp. Match | Adsorbate Type | S. Val | Ads. Match | Comp. Match |
|---|---|---|---|---|---|---|---|
| $*OHCH_2CH_3$ | 0.9733 | 1.0000 | 0.9521 | $*ONN(CH_3)_2$ | 0.9867 | 1.0000 | 0.9662 |
| *H | 0.9800 | 1.0000 | 0.9320 | *CH*CH | 0.9667 | 1.0000 | 0.9862 |
| *COHCOH | 0.9800 | 1.0000 | 0.9796 | $*OCHCH_3$ | 0.9800 | 1.0000 | 0.9592 |
| $*NO_2NO_2$ | 0.9733 | 1.0000 | 0.9658 | $*CH_2*CO$ | 0.9800 | 1.0000 | 0.9728 |
| $*CH_4$ | 0.9733 | 1.0000 | 0.9658 | *NHNH | 0.9667 | 1.0000 | 0.9448 |
| *CCH | 0.9933 | 1.0000 | 0.9597 | $*CH_2CH_2OH$ | 0.9867 | 1.0000 | 0.9324 |
| $*CHOHCH_3$ | 0.9800 | 1.0000 | 0.9320 | $*CHCH_2OH$ | 0.9800 | 1.0000 | 0.9524 |
| $*ONNH_2$ | 0.9867 | 1.0000 | 0.9865 | *C*C | 0.9867 | 1.0000 | 0.9459 |
| $*CCH_2OH$ | 0.9933 | 1.0000 | 0.9664 | *CCO | 0.9933 | 1.0000 | 0.9530 |
| *CN | 0.9800 | 1.0000 | 0.9796 | *CHOCHOH | 1.0000 | 1.0000 | 0.9533 |
| $*NO_3$ | 0.9867 | 1.0000 | 0.9797 | $*OHNH_2$ | 0.9600 | 1.0000 | 0.9583 |
| *N*NH | 1.0000 | 1.0000 | 0.9467 | $*CH_2CH_3$ | 0.9867 | 1.0000 | 0.9865 |
| $*COHCH_3$ | 0.9867 | 1.0000 | 0.9662 | *C | 0.9867 | 1.0000 | 0.9122 |
| $*NH_3$ | 0.9667 | 1.0000 | 0.9379 | $*CCH_2$ | 0.9933 | 1.0000 | 0.9664 |
| *NH | 0.9733 | 1.0000 | 0.9315 | *CHCHO | 0.9933 | 1.0000 | 0.9530 |
| *CH*COH | 0.9867 | 1.0000 | 0.9527 | *CHOH | 0.9800 | 1.0000 | 0.9660 |
| *CHOHCHOH | 1.0000 | 1.0000 | 0.9800 | *COHCHOH | 0.9800 | 1.0000 | 0.9796 |
| $*CHCH_2$ | 0.9667 | 1.0000 | 0.8966 | $*CHOCH_2OH$ | 0.9800 | 1.0000 | 0.9592 |
| *CCHO | 0.9933 | 1.0000 | 0.9463 | *N | 0.9867 | 1.0000 | 0.9054 |
| *N*NO | 0.9733 | 1.0000 | 0.9452 | *OH | 0.9867 | 1.0000 | 0.9527 |
| $*COCH_2O$ | 0.9867 | 1.0000 | 0.9797 | $*COHCH_3$ | 1.0000 | 1.0000 | 0.9600 |
| $*CCH_3$ | 0.9933 | 1.0000 | 0.9128 | $*OHCH_3$ | 0.96000 | 1.0000 | 0.9514 |
| $*CH_2$ | 0.9733 | 1.0000 | 0.9452 | $*CH_3$ | 0.9733 | 1.0000 | 0.9247 |
| $*CH_2O$ | 0.9800 | 1.0000 | 0.9048 | $*OCH_2CH_3$ | 0.9933 | 1.0000 | 0.9664 |
| $*CH_2OH$ | 0.9933 | 1.0000 | 0.9396 | *CCHOH | 0.9800 | 1.0000 | 0.9864 |
| $*OCH_3$ | 0.9933 | 1.0000 | 0.9060 | *CHO*CHO | 0.9733 | 1.0000 | 0.9589 |
| $*CHOHCH_2OH$ | 0.9800 | 1.0000 | 0.9252 | *CHCHOH | 0.9867 | 1.0000 | 0.9527 |
| $*OCH_2CHOH$ | 0.9800 | 1.0000 | 0.9728 | $*OH_2$ | 0.9800 | 1.0000 | 0.9456 |
| $*N_2$ | 0.9933 | 1.0000 | 0.9732 | $*COHCH_2OH$ | 0.9733 | 1.0000 | 0.9589 |
| *COCHO | 0.9800 | 1.0000 | 0.9456 | $*NO_2$ | 0.9867 | 1.0000 | 0.9392 |
| *O | 0.9733 | 1.0000 | 0.9247 | *CHCO | 0.9800 | 1.0000 | 0.9660 |
| *COHCHO | 0.9800 | 1.0000 | 0.9524 | *CHOHC$H_2$ | 0.9867 | 1.0000 | 0.9662 |
| *ONH | 0.9867 | 1.0000 | 0.9595 | *NONH | 0.9733 | 1.0000 | 0.9795 |
| *NO | 0.9800 | 1.0000 | 0.9320 | | | | |